\documentclass[letterpaper]{article}
\usepackage[preprint]{amlc_2022}

\usepackage{times}

\usepackage{mathtools}
\usepackage{graphicx}
\usepackage{mathrsfs}
\usepackage{amsmath}
\usepackage{amssymb}
\usepackage{natbib}
\usepackage{soul,color}
\usepackage{algorithm,algcompatible}
\usepackage{tabularx}
\usepackage{mathrsfs}
\usepackage{multirow}
\usepackage{bbm}
\usepackage{amsthm}
\usepackage{bbold}
\usepackage{color,soul}

\pdfoutput=1
\usepackage{hyperref}
\hypersetup{
  pdfinfo={
    Title={title},
    Author={author},
  }
}
\usepackage{pdfpages}

\title{Contact Complexity in Customer Service}

\author{ {\bf Shu-Ting~Pi} \\
Amazon   \\
Cupertino, CA 95014\\
shutingp@amazon.com\\
\And
{\bf Michael Yang}   \\
Amazon \\
Seattle, WA 98109    \\
abyang@amazon.com
\And
{\bf Qun Liu}  \\
Amazon           \\
Seattle, WA 98109 \\
qunlin@amazon.com
}

\begin{document}

\maketitle






\title{Contact Complexity in Customer Service}






\begin{abstract}
Customers who reach out for customer service support may face a range of issues that vary in complexity. Routing high-complexity contacts to junior agents can lead to multiple transfers or repeated contacts, while directing low-complexity contacts to senior agents can strain their capacity to assist customers who need professional help. To tackle this, a machine learning model that accurately predicts the complexity of customer issues is highly desirable. However, defining the complexity of a contact is a difficult task as it is a highly abstract concept. While consensus-based data annotation by experienced agents is a possible solution, it is time-consuming and costly. To overcome these challenges, we have developed a novel machine learning approach to define contact complexity. Instead of relying on human annotation, we trained an AI expert model to mimic the behavior of agents and evaluate each contact's complexity based on how the AI expert responds. If the AI expert is uncertain or lacks the skills to comprehend the contact transcript, it is considered a high-complexity contact. Our method has proven to be reliable, scalable, and cost-effective based on the collected data. 

\end{abstract}
\maketitle
\section{Introduction}
E-commerce customers may require varying levels of support when contacting customer service for an issue. A complex problem might require assistance from a senior agent, while a simple issue, such as a return or refund, can be resolved by a junior agent. Incorrectly routing a customer with a complex issue to a junior agent can result in poor customer experiences, such as transfers or repeated contacts. Conversely, routing a customer with a simple issue to a senior agent is costly for the business and limits the availability of senior agents for complex problems.

Traditionally, contacts are routed through a machine learning model based on product line or service type. However, many contacts do not fit this model well and customers can experience different levels of complexity within a single product family. For example, a refund for a smartphone is simpler to resolve than a connectivity issue that requires troubleshooting. Additionally, customers may encounter problems that span multiple products or services.

The aim of routing is to provide customers with the best possible solutions, resulting in an optimal customer experience at the lowest cost for the business. To achieve this, we propose a new routing mechanism based on contact complexity. Upon reaching customer service, the customer's profile will be evaluated by a complexity model to determine its complexity score. If the score is high, the contact will be routed directly to a senior agent. If the score is low, it will be routed to a junior agent. Contacts with a medium complexity score will be sent to the standard product-based routing model. The addition of the complexity dimension in the routing workflow can significantly reduce contact transfers, repeat contacts, and unnecessary costs. A workflow chart of our proposed routing pipeline can be found in Appendix Fig. 1.

However, defining "complexity" has proven to be a challenging task. This is due to several reasons, such as the highly subjective nature of complexity where a contact that is difficult for a junior agent may be easy for a senior agent. Additionally, only senior agents are capable of accurately determining the complexity of a contact. While consensus-based data annotation by senior agents could provide reliable and professional labels, the process of having them label thousands, if not millions, of contacts to create a dataset is slow and expensive.

To address these challenges, we have developed a novel machine learning approach to define contact complexity. Since senior agents are rare and costly, we trained an AI expert model that mimics their behavior. The complexity of a contact is evaluated based on how the AI expert reacts to it. This results in a well-defined score, ranging from 0 to 1, which can be used as the ground truth label for training downstream routing models. Our method has been validated using objective and subjective metrics and proven to be scalable and cost-effective as it does not require human annotation.

\textbf{Our Contributions} 
The key contributions of our work are:
\begin{itemize}
    \item We propose an efficient and scalable method to determine contact complexity without the need for human annotation.
    \item Our approach not only provides a complexity score but also generates other relevant features such as sentence length, entropy, and skillfulness. These features capture the high-level content of a transcript and can be used as input for downstream machine learning models.
    \item Our work highlights the value of the intermediate steps of an ensemble boosting model. This aspect of the model is often overlooked in the scientific community, but our findings demonstrate that it can provide valuable insights into the input data.
\end{itemize}

\textbf{Related Works} 
Our work draws significant parallels with previous studies that employ information theory quantities to characterize input data properties. The information bottleneck concept, for instance, implements similar approaches and calculates mutual information between intermediate and final layer activations in a neural network to illustrate how data point noise is filtered through the network \citep{bottleneck1,bottleneck2}. While our method is similar, we use KL divergence and ensemble boosting.

Another area closely related to our project is the classification of text difficulty, which examines the readability of articles for different age groups of children to aid educators in crafting appropriate materials. Our model is a prime example of how text difficulty can be linked to customer service. Other related studies on machine learning applications in this field can be found in \citep{difficulty1} and \citep{difficulty2}. Additionally, the implementation of text difficulty in business settings such as healthcare is a noteworthy topic \citep{difficulty3}.



\begin{figure*}
  \centering
  \includegraphics[width=1.0\textwidth]{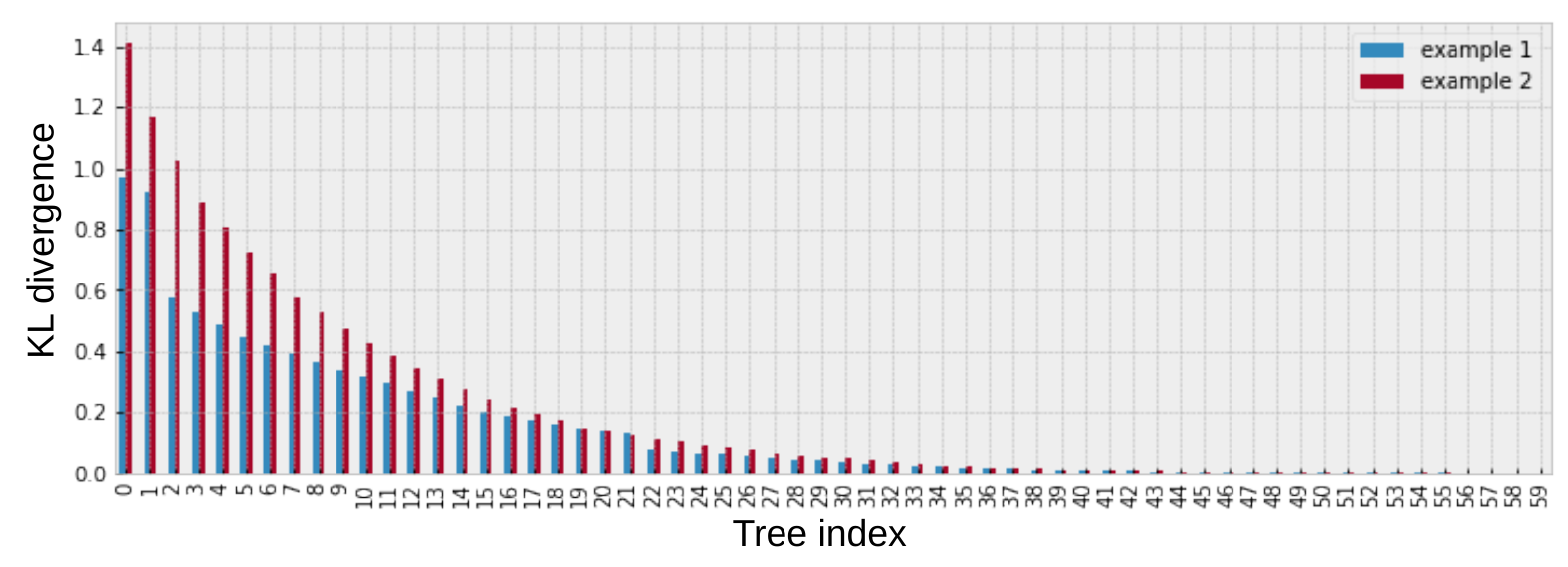}
  \caption{The KL divergence boosting function is demonstrated for two different examples. The boosting model has 60 trees. The divergence boosting function of Example 1 decays faster compared to Example 2, resulting in a smaller integral, or skillfulness, for Example 1. It is important to note that the divergence boosting always reaches zero when the tree index reaches its maximum number.}\label{boosting_function}
\end{figure*}

\section{Metohds}

\subsection{The Dataset}
We utilize a dataset obtained from the MessageUS channel at Amazon. MessageUS is a text-based platform that allows customers to communicate with Amazon's customer service agents. The dataset includes multiple features for each customer interaction, such as chat transcripts, handling time, the requirement for third-party assistance, and the resolution of the customer's issue.

After each dialogue session, agents manually select a standardized issue code (SIC) to summarize the topic and proposed solutions of each customer interaction. This SIC system is not unique to Amazon, as most eCommerce companies have their own version to provide a concise summary of the conversation.

It's crucial to note that the dataset only contains customer text data, and all confidential information, such as names and account details, has been masked to protect privacy before being made accessible to researchers. Although the dataset is obtained from Amazon's internal database, its format is standardized and comparable to customer service data that most companies possess. As a result, the methodology presented in this article is applicable to any other e-commerce company.

\subsection{The AI expert}
An AI expert is a machine learning model that has domain knowledge of the routing business. Thus, the simplest approach is to train a classifier that predicts the corresponding SIC code for a given contact transcript.

To train the model, we used 450,000 contact transcripts from January to March 2021 as input and aimed to predict one of the 152 possible SIC codes. The model algorithm was based on TF-IDF text embedding\citep{tf-idf} and gradient boosting trees\citep{gbdt}, powered by LightGBM\citep{LightGBM}. Similar algorithms, such as Xgboost\citep{xgboost} or Catboost\citep{catboost}, could also be utilized. The model performance on the validation set was impressive, with an AUC$_{\mu}$\citep{AUC_mu} score of 0.96 and top-15 accuracy of 0.93, indicating strong connections between the transcripts and the SIC codes.

Once the classifier is obtained, it can be considered an "AI Expert" with the domain knowledge to predict SIC codes. However, training a classifier for SIC codes is just one option. Another approach is to design a multi-label classifier that predicts multiple labels relevant to the routing business, such as the need for third-party assistance, the resolution of the issue, and the requirement for a transfer. This could be accomplished with shared-weights neural networks\citep{deeplearning} or multi-label ensemble trees\citep{RandomForest}. This approach would offer a more knowledgeable AI Expert that can better identify the complexity of customer service issues. In this article, we only use the SIC code as a proof of concept.

\subsection{Hypotheses of Complexity}
We believe the following attributes can characterize a high-complexity contact:

\begin{itemize}
    \item \textbf{Length}: A high-complexity contact will need multiple steps to diagnose the issues. As a result, their transcripts are usually longer than others.
    \item \textbf{Uncertainty}: A high-complexity contact is more difficult to understand. As a result, agents will be more uncertain about the issues that customers are facing. 
    \item \textbf{Skillfulness}: A high-complexity contact should require more skills to resolve the issue. Only experienced and skillful agents can adequately handle it. 
\end{itemize}

In short, we hypothesize that a high-complexity contact should be lengthy, highly uncertain, and require more skills. In contrast, a low-complexity contact is short, highly confident, and requires fewer skills. If so, our task becomes clear: how can we convert the above hypotheses from concepts into numbers?

\subsection{Hypotheses to Mathematics}
We use the following three quantities to represent the length, uncertainty, and skillfulness:

\begin{itemize}
    \item \textbf{length ($\mathcal{L}$) = number of agent’s sentences}
    We define the number of agent's sentences in a transcript as the length of a contract:
    \[\mathcal{L} = \sum_{i}\delta_{S_{i}, agent}\]
    , where $\delta$ is kronecker delta function, $S_{i}$ is the speaker of $i$-th sentence in a contact transcript. The above function is essential to count the sentences that agents speak. We intentionally ignore customers' sentences because a customer may have multiple sentences of conversation with the chatbot, and these sentences are usually less informative.
    \item \textbf{uncertainty ($\mathcal{E}$) = the entropy of AI expert's output} When a transcript is fed to our AI expert (a classifier that predicts the SIC code), the expert will output the predicted probability on each SIC code. If the expert is pretty confident about its prediction, the entropy of the output probability will be low and vice versa. Therefore, we can use entropy as an indicator of uncertainty \citep{info_entropy}:
    \[\mathcal{E} = -\sum_{i}P_{i}log(P_{i}) \]
    , where $P_{i}$ is the probability on class $i$.
    \item \textbf{skillfulness ($\mathcal{S}$) = the integral of divergence boosting function} 
    The gradient boosting tree model combines the predictions of multiple ensemble trees to produce the final prediction. The more trees used, the more skillful the model becomes. The difference between the outputs of a weak model (fewer trees) and a strong model (more trees) can be evaluated using KL divergence, which measures the difference between two probability distributions\citep{kl_divergence}. The boosting function \citep{boosting} of KL divergence is defined as $\phi(i)=D_{KL}(P_{i}||P_{M})$, $i = 1 \sim M$, where $P_{i}$ represents the output distribution using the first $i$ trees, and $M$ represents the total number of trees in the model.

    The integral of the boosting function
    \[\mathcal{S}=\sum_{i}^{M}\phi(i)=\sum_{i}^{M}D_{KL}(P_{i}||P_{M}) \]
    can be considered an indicator of the model's skillfulness, with a slow decay to zero indicating the need for more knowledge or skills, and a quick decay to zero indicating less need. Figure 1 shows two examples of how the divergence boosting function behaves. The boosting function of the first example decreases faster than the second, indicating that the model requires more skills to handle the second example properly.

\end{itemize}

\subsection{Distributions of Hypotheses} 
To simplify the characterization of complexity, we have defined three attributes: length, entropy, and skillfulness. These attributes were calculated for 450K contacts used to train the AI expert, and the results are shown in Figures 2(a) to (c). The distributions of the attributes have long-tailed structures, with high-complexity contacts lying in the tail (red) region and low-complexity contacts lying in the head (green) region.

However, using three separate numbers to describe complexity can be complicated for the routing business. Therefore, we aim to use a single variable that represents complexity in a more straightforward manner. To do this, we normalize all the attributes to a similar range and combine them with appropriate weights to create a single variable.

\begin{figure*}
  \centering
  \includegraphics[width=1.0\textwidth]{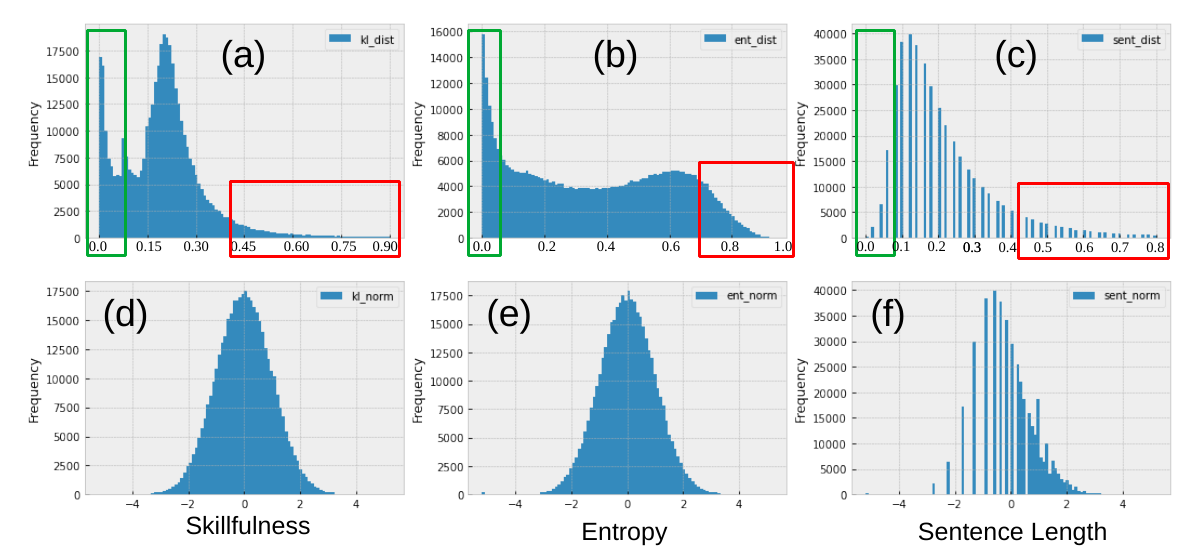}
  \caption{(a)-(c) show the distribution of 450K training contacts for different hypotheses: (a) $\mathcal{S}$, (b) $\mathcal{E}$, and (c) $\mathcal{L}$. (Note: To protect Amazon's business from potential information leaks, we rescaled $\mathcal{S}$, $\mathcal{E}$ and $\mathcal{L}$ between 0 and 1. However, this rescaling does not affect the findings presented in this article since only the relative values are significant in the subsequent distribution transformations.) (d)-(f) show the distribution of the 450K training contacts after quantile transformation, which converts all hypotheses to a normal distribution. The transformed hypotheses are referred to as (d) $\mathcal{S}^N$, (e) $\mathcal{E}^N$, (f) $\mathcal{L}^N$, respectively.}\label{flow_chart}
\end{figure*}

\subsection{Complexity Score}
The differences in scale and shape of the distributions make it risky to merge them into one variable, as it could skew the significance of individual hypotheses. To obtain a single variable $\mathcal{C}$, we perform a quantile transformation to normalize the distributions (mean=0, variance=1), as depicted in Fig.2(d)-(f). This way, if a contact has high/low values on all three hypotheses, $\mathcal{C}$ will retain its high/low value. The transformed hypotheses are then safely combined.

The quantile transformed hypotheses, represented as $\mathcal{L}^{N}$, $\mathcal{E}^{N}$, and $\mathcal{S}^{N}$, are combined as $\mathcal{C}=w\times\mathcal{L}^{N}+\mathcal{E}^{N}+\mathcal{S}^{N}$ with the introduction of a parameter $w$ for sentence length. This is intentional for two reasons: (1) contacts with short transcripts are less likely high-complexity contacts despite having high entropy and skillfulness, and increasing the weight on sentence length can improve precision in identifying high-complexity contacts; and (2) adding the normalized hypotheses together ($w=1$) results in a negatively skewed distribution, as seen in Fig.3(a). However, increasing the weight on $\mathcal{L}$ can tune the distribution towards a symmetrical shape, as demonstrated in Fig.3(a)-(c). Although symmetry is not a requirement, having a distribution where extreme simplicity or complexity is rare, with most contacts in the medium region, is desirable. Thus, we choose to use $w=2$ to define $\mathcal{C}$, resulting in a Gaussian distribution with desirable properties.

Although the complexity score $\mathcal{C}$ accurately reflects the complexity, it may not be easily understood by non-technical people. To address this, we perform a quantile transformation to convert $\mathcal{C}$ from a Gaussian distribution ($w=2$) to a uniform distribution, as shown in Fig.3(d). To distinguish the two, we refer to $\mathcal{C}$ as the "absolute complexity score" (Fig.3(c)) and the transformed score, $\mathcal{Q} = \mathcal{T^U_G}\mathcal{C}$, as the "relative complexity score" (Fig.3(d)). The transformation, $\mathcal{T^U_G}$, converts $\mathcal{C}$ (Gaussian distribution) to $\mathcal{Q}$ (uniform distribution), and is calculated as follows:
\[\mathcal{Q} = \mathcal{T^U_G C} = \mathcal{T^U_G}(2\times\mathcal{L+E+S})\]

The definition of $\mathcal{Q}$ is unambiguous. Our dataset comprises 450K contacts and we calculate the complexity score $\mathcal{C}$ for each one. If a contact's absolute complexity score $\mathcal{C}$ is above 95\% of all scores, its relative complexity score $\mathcal{Q}$ will be 0.95. On the other hand, if a contact's absolute complexity score is only above 5\% of all scores, its relative complexity score will be 0.05.

Since $\mathcal{Q}$ is a value that ranges between 0 and 1 with a straightforward interpretation, we will refer to it as the "complexity score" in subsequent discussions unless stated otherwise. This is the score we will use in our further analysis.

\begin{figure*}
  \centering
  \includegraphics[width=1.0\textwidth]{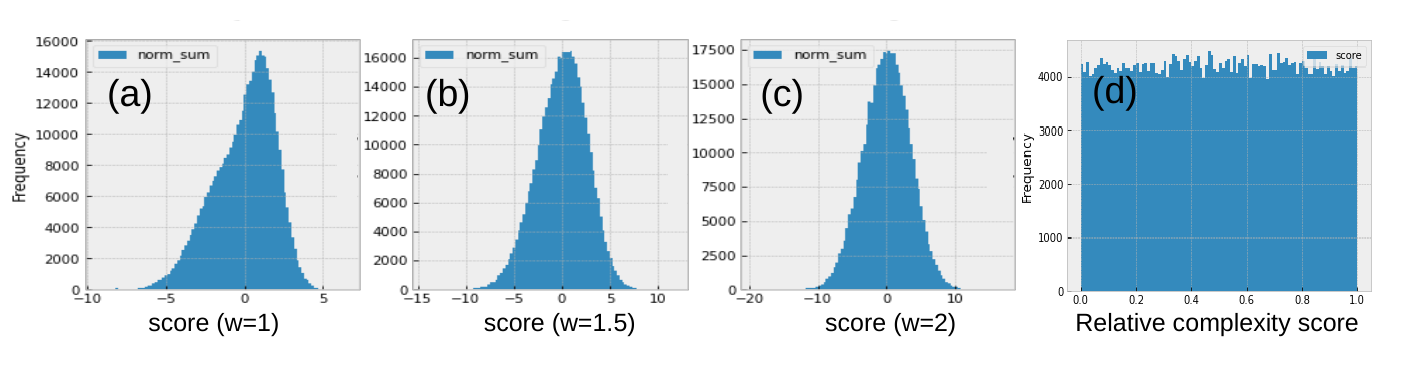}
  \caption{(a)-(c) show the distribution of the absolute complexity score for different weights $w$ on $\mathcal{L}^N$. As $w$ increases, the distribution changes from negatively skewed to symmetrical. (d) shows the relative complexity score, obtained by performing a quantile transform on (c) to map it to a uniform distribution. }\label{uniform_modify}
\end{figure*}

\section{Results}
This section will discuss whether the complexity score can reflect a contact's complexity.  
\subsection{Extreme Cases}
We have a goal of routing simple contacts to junior agents and complex contacts to senior agents. For this reason, we are particularly interested in cases where the complexity score, $\mathcal{Q}$, is extremely low or high. In Fig. 4, we present the distributions of all hypotheses, i.e. $\mathcal{L}$, $\mathcal{E}$, and $\mathcal{S}$, for contacts with $\mathcal{Q} < 0.05$ (pink), $0.05 < \mathcal{Q} < 0.95$ (green), and $\mathcal{Q} > 0.95$ (blue), respectively. As seen in the figure, low-complexity contacts have relatively low values on all hypotheses, whereas high-complexity contacts tend to have similar properties but in the opposite direction. Contacts with medium complexity scores have mixed properties in their distributions. This demonstrates that the complexity score, $\mathcal{Q}$, effectively represents the level of complexity of a contact across all hypotheses. 
\begin{figure*}
  \centering
  \includegraphics[width=1.0\textwidth]{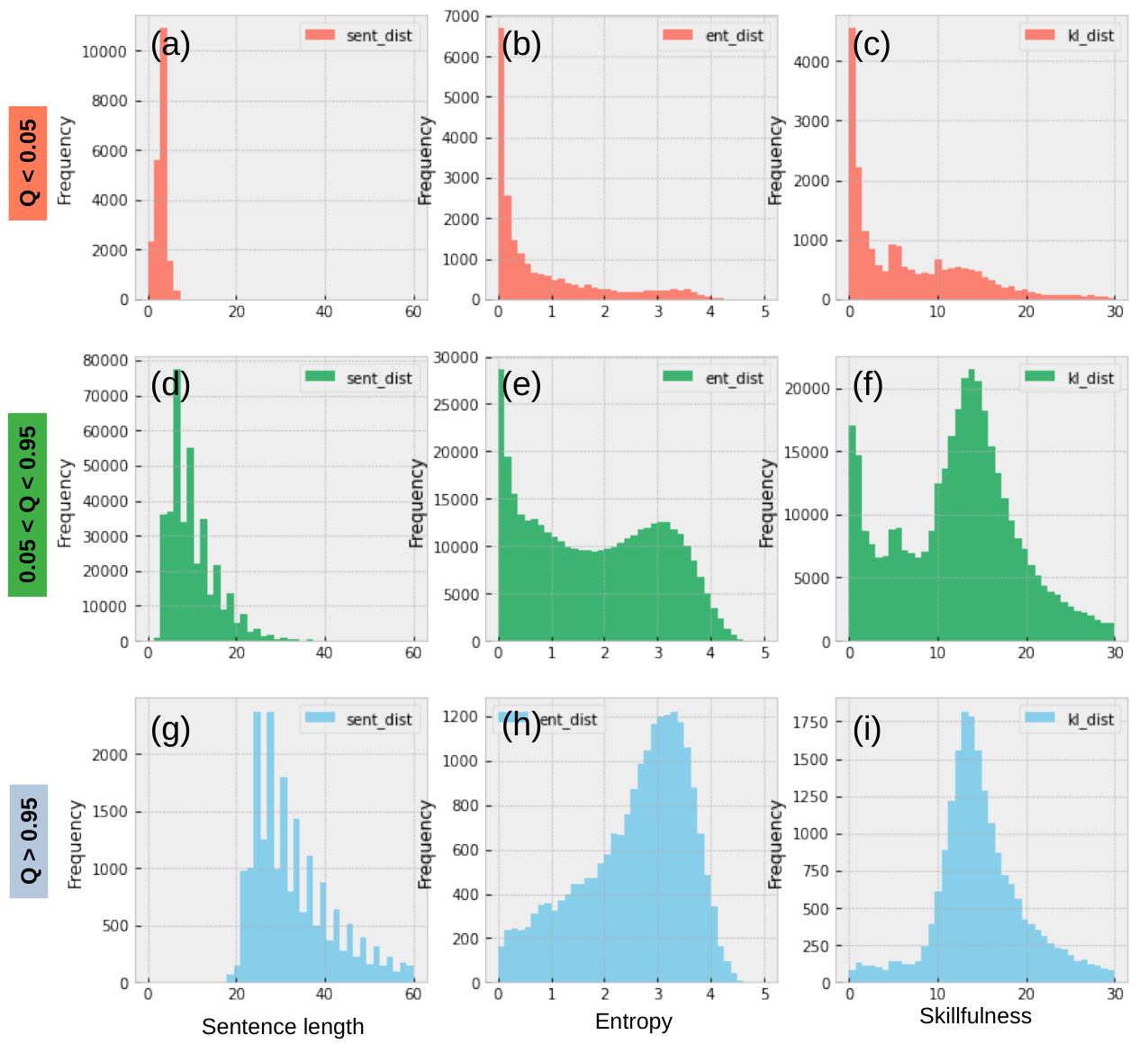}
  \caption{The distributions of hypotheses $\mathcal{L}$, $\mathcal{E}$ and $\mathcal{S}$ are presented for high, medium, and low complexity contacts. (a)-(c) represent low-complexity contacts ($\mathcal{Q}<0.05$), (d)-(f) represent medium-complexity contacts ($0.05 \leq \mathcal{Q} \leq 0.95$), and (g)-(i) represent high-complexity contacts ($\mathcal{Q}>0.95$). Low-complexity contacts tend to have small values on all hypotheses, while high-complexity contacts have similar behaviors but opposite directions.}\label{extreme_case}
\end{figure*}
\subsection{Indirect Evidences}
So far, there is no established metric to evaluate the complexity of a real-world contact. However, we have some indirect evidence to assess the validity of our approach to define contact complexity.

We conducted a visual inspection of 400 examples, which includes 200 high-complexity contacts with a complexity score $\mathcal{Q}>0.95$ and 200 low-complexity contacts with a complexity score $\mathcal{Q}<0.05$. The results showed that high-complexity contacts had a low resolution rate (resolved contacts/200) and a high transfer rate (transferred contacts/200). In contrast, low-complexity contacts had a high resolution rate and a low transfer rate. According to our survey, only 26\% of high-complexity contacts were resolved, and 62\% underwent at least one transfer. On the other hand, 85\% of low-complexity contacts were resolved, and only 11\% underwent at least one transfer. (Note: The resolution or unresolved status was determined by reviewing the transcript and determining if the customer's issue was resolved during the session. Ticket cutting, third-party assistance requests, or follow-up calls were not considered resolved.) Although the dataset is limited and the resolution/transfer of a contact may not be solely due to its complexity, the significant difference between the two groups indicates that our complexity score accurately captures the concept of complexity.

\begin{figure*}
  \centering
  \includegraphics[width=1.0\textwidth]{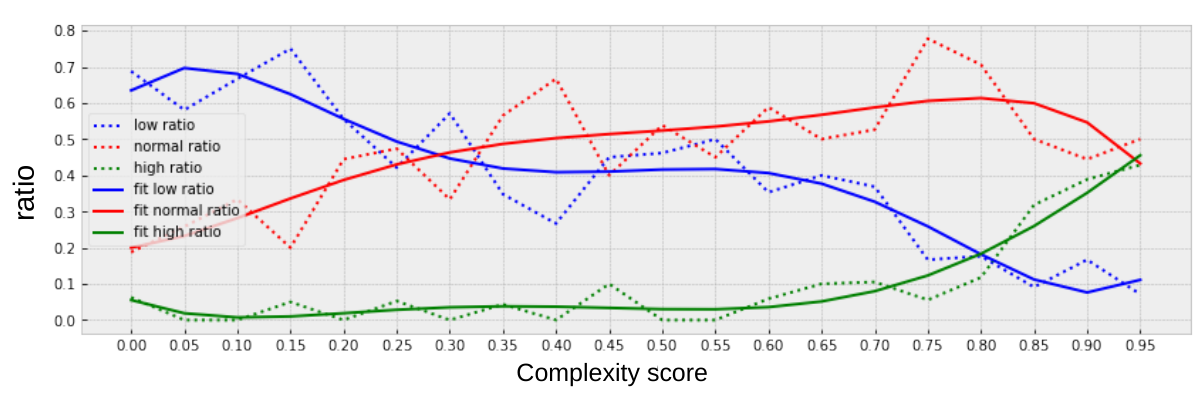}
  \caption{This plot shows the probability of finding high, medium, or low complexity based on different scores. Over 400 ground truth labels were generated by senior agents. The dash lines represent the actual numbers, while the solid lines are fitted curves using polynomial functions. The blue line represents low complexity, the red line represents medium complexity, and the green line represents high complexity.}\label{info_score}
\end{figure*}

\subsection{Direct Evidence}
We have validated our approach to defining contact complexity through both indirect and direct evidence. We collaborated with senior Amazon agents to label the complexity of 400 examples, ensuring at least two agents reviewed each contact and reached a consensus label of "low complexity," "normal," or "high complexity." Fig.5 shows the results of binning the complexity scores into 20 intervals and counting the probability of finding a low, normal, or high complexity example in each interval. The likelihood of finding a low complexity contact decreases gradually to 0.1 as the complexity score increases, while the possibility of finding a high complexity contact rapidly increases to 0.5 after a complexity score of 0.6. Our visual inspection of contacts labeled as "normal" with $\mathcal{Q} > 0.9$ confirmed that they were relatively tricky but not considered high complexity by the agents. This combined direct and indirect evidence supports the consistency of our complexity score with the agents' understanding of contact complexity.

\section{Conclusion}
We have developed a unique method to measure contact complexity using an AI expert. Our approach involves training the AI expert to mimic the process of assigning SIC codes to customer contacts and evaluating contact complexity based on the AI expert's response. A contact is considered high complexity if it is lengthy, uncertain, and requires more skills, and low complexity if it is short, straightforward, and requires fewer skills. Our approach has been validated through both indirect and direct methods and has been found to be highly consistent with human perceptions of complexity.

With these results, we can train a routing model based on pre-contact data to route high and low-complexity contacts to appropriate agents using the complexity scores as labels. The numbers generated by our models, including the hypotheses, absolute complexity scores, and relative complexity scores, provide valuable information about customer contacts and can be used as features for further analysis.

In the future, we plan to incorporate more automatically generated service labels into our AI expert to further increase its skillfulness and knowledge. Additionally, we believe that using an ensemble of AI experts to calculate the average complexity score would be an exciting area to explore. An ensemble model would offer a more robust and stable measurement of contact complexity and provide better performance in real-world applications.

\bibliographystyle{apalike}
\bibliography{citation.bib}

\begin{thebibliography}{}

\bibitem[Balyan et~al., 2020]{difficulty1}
Balyan, R., McCarthy, K.~S., and McNamara, D.~S. (2020).
\newblock Applying natural language processing and hierarchical machine
  learning approaches to text difficulty classification.
\newblock {\em International Journal of Artificial Intelligence in Education
  volume}, 30:337--370.

\bibitem[Balyan et~al., 2021]{difficulty2}
Balyan, R., McCarthy, K.~S., and McNamara, D.~S. (2021).
\newblock Comparing machine learning classification approaches for predicting
  expository text difficulty.
\newblock {\em International Florida Artificial Intelligence Research Society
  Conference}.

\bibitem[Chen and Guestrin, 2016]{xgboost}
Chen, T. and Guestrin, C. (2016).
\newblock Xgboost: A scalable tree boosting system.
\newblock {\em ACM SIGKDD International Conference}.

\bibitem[Friedman, 2001]{gbdt}
Friedman, J.~H. (2001).
\newblock Greedy function approximation: a gradient boosting machine.
\newblock {\em Annals of statistics}, pages 1189--1232.

\bibitem[Goodfellow et~al., 2016]{deeplearning}
Goodfellow, I., Bengio, Y., and Courville, A. (2016).
\newblock Deep learning.
\newblock {\em The MIT Press}.

\bibitem[Ho, 1995]{RandomForest}
Ho, T.~K. (1995).
\newblock Random decision forests.
\newblock {\em Proceedings of the 3rd International Conference on Document
  Analysis and Recognition}, pages 278--282.

\bibitem[Ke et~al., 2017]{LightGBM}
Ke, G., Meng, Q., Finley, T., Wang, T., Chen, W., Ma, W., Ye, Q., and Liu,
  T.-Y. (2017).
\newblock Lightgbm: A highly efficient gradient boosting decision tree.
\newblock {\em Neural Information Processing Systems (NIPS)}.

\bibitem[Kleiman and Page, 2019]{AUC_mu}
Kleiman, R.~S. and Page, D. (2019).
\newblock Auc$_{\mu}$: A performance metric for multi-class machine learning
  models.
\newblock {\em International Conference on Machine Learning (ICML)}, PMLR(97).

\bibitem[Kullback and Leibler, 1951]{kl_divergence}
Kullback, S. and Leibler, R. (1951).
\newblock On information and sufficiency.
\newblock {\em Annals of Mathematical Statistics}, 22:79--86.

\bibitem[Prokhorenkova et~al., 2018]{catboost}
Prokhorenkova, L., Gusev, G., Vorobev, A., Dorogush, A.~V., and Gulin, A.
  (2018).
\newblock Catboost: unbiased boosting with categorical features.
\newblock {\em d Conference on Neural Information Processing Systems
  (NeurIPS)}.

\bibitem[Rajaraman and Ullman, 2011]{tf-idf}
Rajaraman, A. and Ullman, J.~D. (2011).
\newblock Mining of massive datasets.
\newblock {\em Cambridge University Press}, pages 1--17.

\bibitem[Shannon, 1948]{info_entropy}
Shannon, C.~E. (1948).
\newblock A mathematical theory of communication.
\newblock {\em Bell System Technical Journal}, 27:379--423.

\bibitem[Tishby et~al., 2000]{bottleneck1}
Tishby, N., Pereira, F.~C., and Bialek, W. (2000).
\newblock The information bottleneck method.
\newblock {\em arXiv}, page physics/0004057.

\bibitem[Tishby and Zaslavsky, 2015]{bottleneck2}
Tishby, N. and Zaslavsky, N. (2015).
\newblock Deep learning and the information bottleneck principle.
\newblock {\em arXiv}, page 1503.02406.

\bibitem[Wang, 2006]{difficulty3}
Wang, Y. (2006).
\newblock Automatic recognition of text difficulty from consumers health
  information.
\newblock {\em Computer-Based Medical Systems, 2006. CBMS 2006. 19th IEEE},
  pages 131--136.

\bibitem[Zhou, 2021]{boosting}
Zhou, Z.-H. (2021).
\newblock Ensemble methods: Foundations and algorithms.
\newblock {\em CRC Press}, page~23.

\end{thebibliography}

\newpage
\appendix
\renewcommand\thefigure{\thesection.\arabic{figure}}   
\setcounter{figure}{0} 
\section*{Appendix}
Figure 1 depicts the proposed routing logic for customer service contacts. The customer's profile is first sent to a complexity routing model to determine the contact's complexity level. If the predicted score is low, the contact will be directed to junior agents for cost-effective handling. Contacts with high scores will be directed to senior agents for advanced assistance, while those with intermediate scores will be processed by a product-line-based routing model.

Figure 2 shows an example of a contact transcript with the sentence describing the contact's topic highlighted. This is classified as a low complexity contact with a complexity score of $\mathcal{Q}<0.05$.

\newpage
\begin{figure*}
  \centering
  \includegraphics[width=0.8\textwidth]{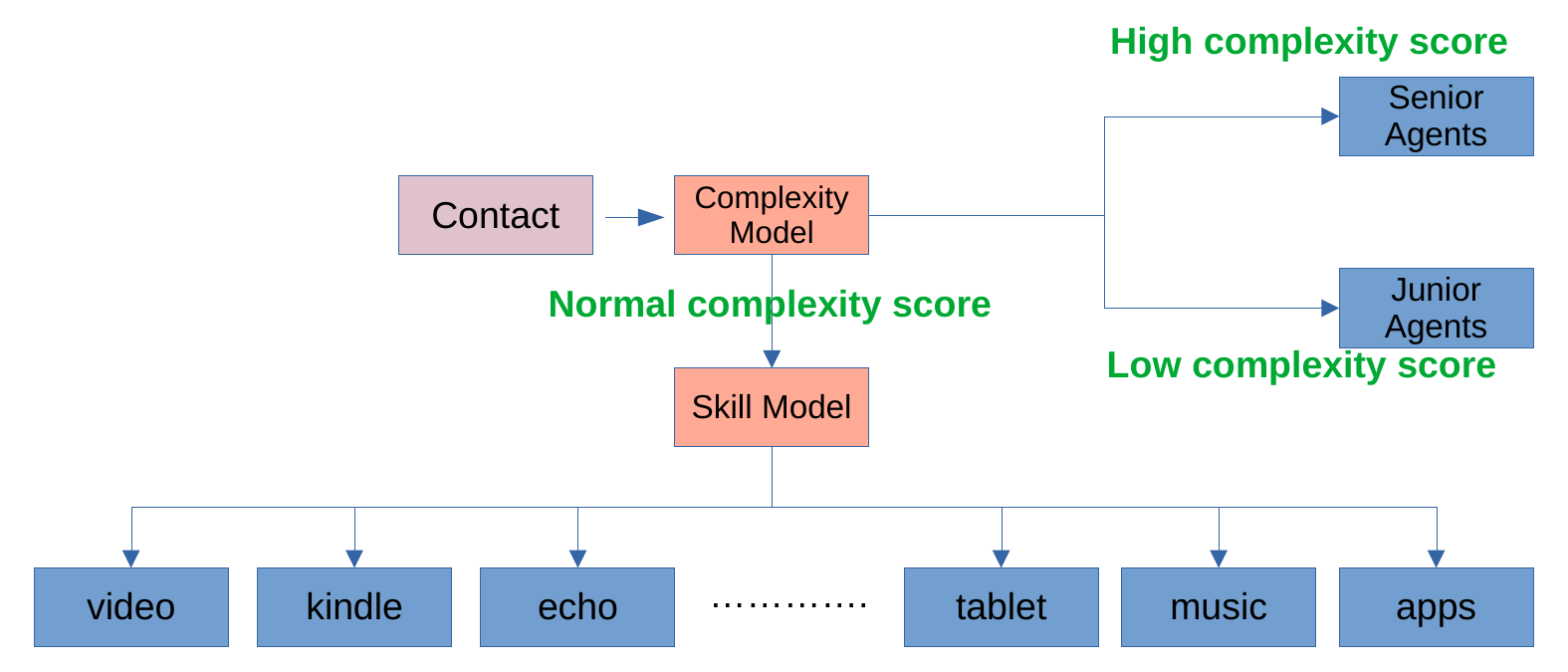}
  \caption{Our proposed routing process involves two steps. Firstly, a complexity model will assess the contact to determine whether it should be routed directly to senior or junior agents. If the contact does not meet the criteria for direct routing, a product-line-based model (here we use Amazon's products to demonstrate the idea) will then analyze the intent of the contact and direct it to the appropriate agents.}\label{pipeline}
\end{figure*}

\begin{figure*}
  \centering
  \includegraphics[width=1.0\textwidth]{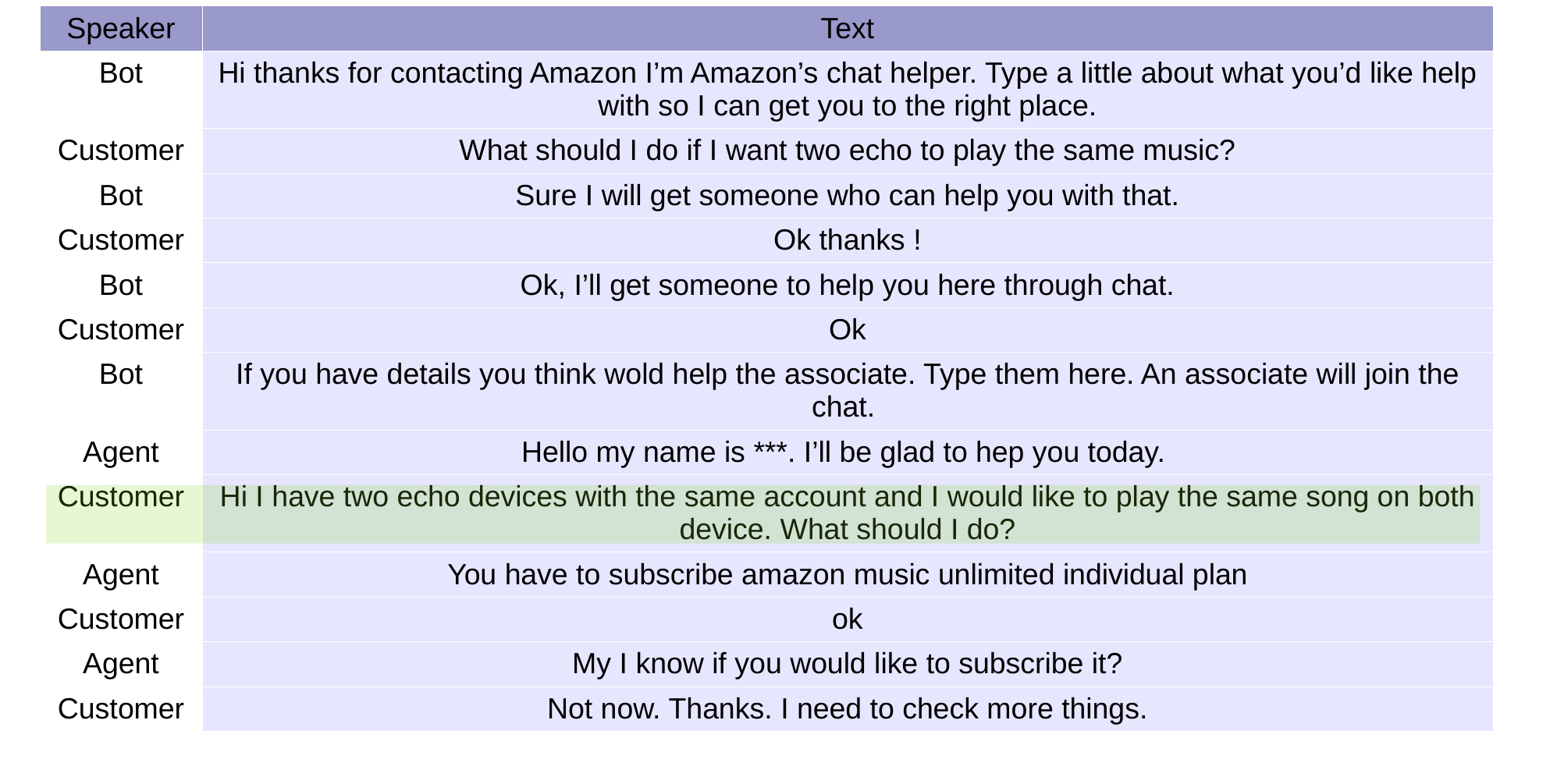}
  \caption{Example of a transcript. The highlighted sentence reflects the topic / customer's issue of this contact.}\label{transcript}
\end{figure*}
\end{document}